\documentclass[journal]{IEEEtran}
\usepackage{cite}

\ifCLASSINFOpdf
   \usepackage[pdftex]{graphicx}
\else
\fi
\usepackage{amsmath}
\ifCLASSOPTIONcompsoc
  \usepackage[caption=false,font=normalsize,labelfont=sf,textfont=sf]{subfig}
\else
  \usepackage[caption=false,font=footnotesize]{subfig}
\fi
\usepackage[acronym]{glossaries}
\usepackage{makecell}
\usepackage{color}

\newacronym{als}{ALS}{alternating least squares}
\newacronym{daq}{DAQ}{data acquisition}
\newacronym{fpga}{FPGA}{field programmable gate array}
\newacronym{ldv}{LDV}{laser Doppler vibrometer}
\newacronym{pcb}{PCB}{printed circuit board}
\newacronym{rms}{RMS}{root mean square}
\newacronym{usb}{USB}{Universal Serial Bus}

\newacronym{dp}{DP}{distal phalanges}
\newacronym{mp}{MP}{middle phalanges}
\newacronym{pp}{PP}{proximal phalanges}
\newacronym{mh}{MH}{metacarpals head}
\newacronym{ms}{MS}{metacarpals shaft}
\newacronym{mb}{MB}{metacarpals base}
\newacronym{c}{C}{carpals}

\newcommand{\pp}{\mathbf{p}}

\begin{document}
%
\title{A Wearable Tactile Sensor Array for Large Area Remote Vibration Sensing in the Hand} 
%
%
%

\author{Yitian~Shao,~\IEEEmembership{Student~Member,~IEEE,}
	Hui~Hu,
	and~Yon~Visell,~\IEEEmembership{Member,~IEEE}
\thanks{Y. Shao, H. Hu and Y. Visell are with the Department of Electrical and Computer Engineering, University of California, Santa Barbara, Santa Barbara, CA, 93106 USA (e-mail: yitianshao@ucsb.edu; michaelhuhui@gmail.com; yonvisell@ece.ucsb.edu)}
\thanks{Manuscript received ?? ??, ??; revised ?? ??, ??.}}

%
%

\markboth{IEEE SENSORS JOURNAL}
{IEEE SENSORS JOURNAL}
%



\maketitle

\begin{abstract}
Tactile sensing is a essential for skilled manipulation and object perception, but existing devices are unable to capture mechanical signals in the full gamut of regimes that are important for human touch sensing, and are unable to emulate the sensing abilities of the human hand.  Recent research reveals that human touch sensing relies on the transmission of mechanical waves throughout tissues of the hand. This provides the hand with remarkable abilities to remotely capture distributed vibration signatures of touch contact. Little engineering attention has been given to important sensory system. Here, we present a wearable device inspired by the anatomy and function of the hand and by human sensory abilities.  The device is based on a 126 channel sensor array capable of capturing high resolution tactile signals during natural manual activities. It employs a network of miniature three-axis sensors mounted on a flexible circuit whose geometry and topology were designed match the anatomy of the hand, permitting data capture during natural interactions, while minimizing artifacts. Each sensor possesses a frequency bandwidth matching the human tactile frequency range. Data is acquired in real time via a custom FPGA and an I$^2$C network. We also present physiologically informed signal processing methods for reconstructing whole hand tactile signals using data from this system. We report experiments that demonstrate the ability of this  system to accurately capture remotely produced whole hand tactile signals during manual interactions.
\end{abstract}


%
\IEEEpeerreviewmaketitle

\section{Introduction}
%
%
%
%
%
\IEEEPARstart{T}{he} human limb is endowed with many thousands of sensory neurons and associated sensory end organs that reach densities exceeding 10/mm$^2$ of skin in the fingertip \cite{jones2006human}. This enables us to perform remarkable perceptual feats, from easily discriminating extremely fine differences in surface texture and materials, to detecting nanometer scale surface features  \cite{manfredi2014natural}.  
These perceptual tasks rely on the low-level transduction of mechanical signals spanning a wide frequency range (from approximately 0 to 1000 Hz) and dynamic range (from 10$^{-3}$ to 10$^{-8}$ m) into electrical spikes that are transmitted to the brain.  This neural information gives rise to  conscious experiences of touching objects and surfaces, and enables the great variety of abilities of the hand, including object grasping, palpation, and manipulation \cite{johansson2009coding}. 

\begin{figure}[h]
	\centering
	\includegraphics[width=83mm]{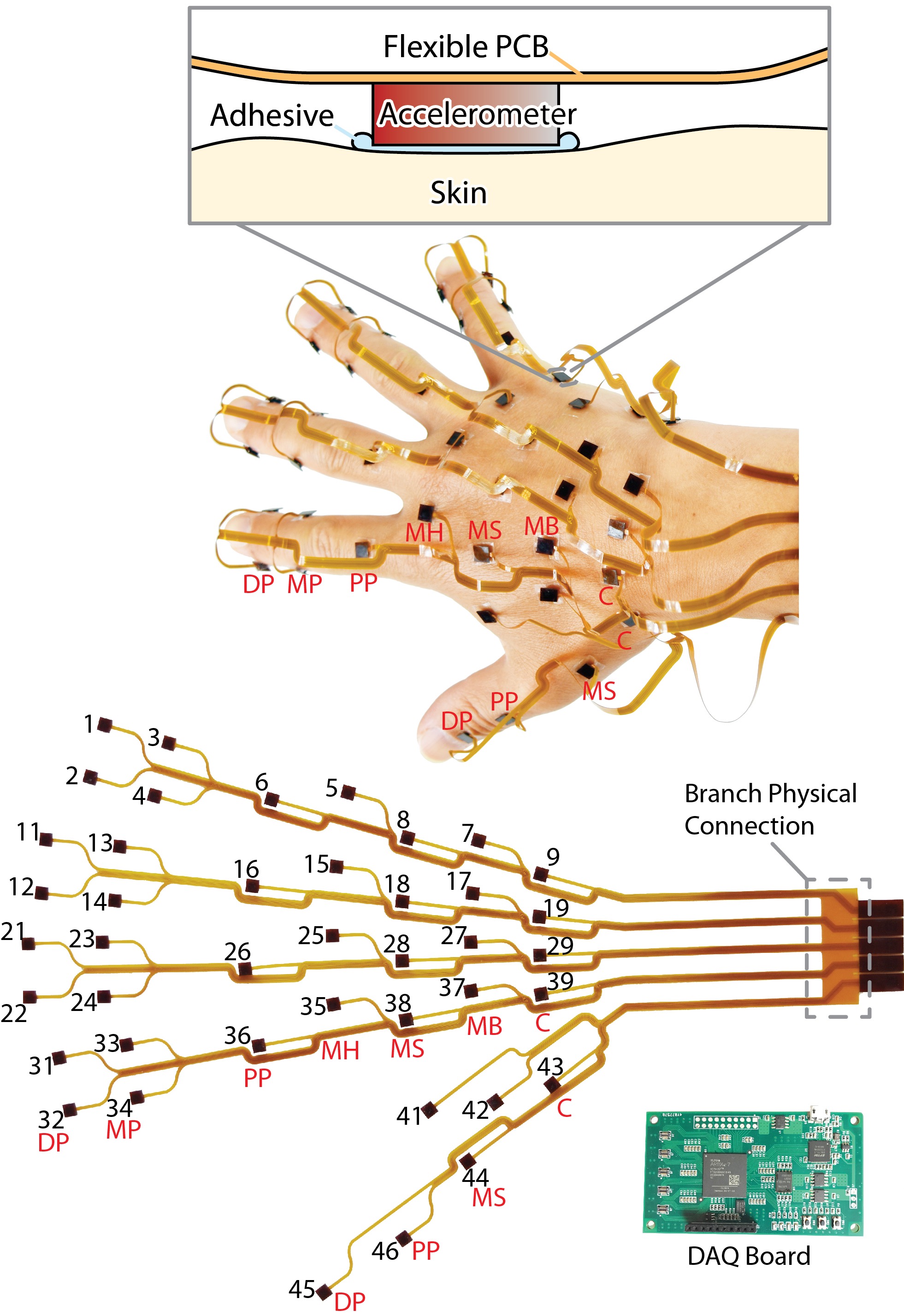}
	\caption{Structure of the wearable sensor and overview of the instrument attached to back of the hand. Each sensor is attached to the skin via a prosthetic adhesive.
	Top view of the sensor array mounted on a flat surface, together with the FPGA-based DAQ board, are shown at the bottom. Each accelerometer is uniquely numbered. Based on the hand anatomy, the array is divided to five branches, in correspondence to digits I to V.
	Sensor locations were specified in relation to anatomical features, on phalanges (DP, MP, PP), metacarpals (MH, MS, MB) and carpals (C), marked in red.
	All five branches have a dedicated electronic connection to the DAQ board.}
	\label{fig:AccIII_Position}
\end{figure}

The human hand is thus a remarkable sensory and prehensile instrument that provides a biological model for engineering systems for robotic manipulation and tactile sensing.  There is growing interest in the engineering of tactile sensors that might be able to reproduce the remarkable feats of perception of biological skin, and that may emulate the large range of motor functions of the human hand. In this research, we present a new wearable tactile sensing array, with which we aim to provide  quantitative information about the tactile signals that are captured by the human hand during natural interactions.  This system also provides a new approach for skin-like artificial  sensing that is inspired by the distributed vibration sensing capacities that are intrinsic to the human hand. It is based on a distributed array of sensors that, in our device, are elastically coupled through the soft tissues of the hand.

\begin{table*}[ht!]
	\renewcommand{\arraystretch}{1.3}
	\caption{Sensing distributed skin vibration: Selected methods described in the literature.}
	\label{tab:Sensor_comparison}
	\centering
	\begin{tabular}{llllll}
		Reference & Sensor Technology & Wearable & Sensor Position & Data Type & Limitations \\
		\hline
		Rossi et al.~\cite{rossi1995hand} & Laser scanning vibrometer & No & Noncontact & Velocity scalar & Requires immobilizing hand\\
		\hline
		Seo et al.~\cite{seo2013estimating} & 3D laser scan  & No & Noncontact & Position vector & Requires immobilization\\
 &&&&& Low frame rate\\
		\hline
		Sakai et al.~\cite{sakai2011vivo} & Optical coherence tomography & No & Noncontact & Tomography & Limited spatial measurement range\\
&&&&& Requires immobilization\\
		\hline
		Tanaka et al.~\cite{tanaka2011noncontact} & High-speed camera & No & Noncontact & Displacement scalar & Requires immobilization\\
		\hline
		Gerhardt et al.~\cite{gerhardt2012novel} & Microscope-video camera & No & Noncontact & Image & Small field of view \\ &&&&& Low frame rate\\
		\hline
		Shirkovskiy et al.~\cite{shirkovskiy2018airborne} & Airborne ultrasound vibrometry & No & Noncontact & Vibrometry image & Requires spatial interpolation \\
&&&&& Requires immobilization \\
		\hline
		Sikdar et al.~\cite{sikdar2007ultrasonic} & Ultrasonic Doppler vibrometry & No & Contact & Vibrometry image & Limited spatial measurement range\\ &&&&& Large sensor mass\\
		\hline
		Sofia et al.~\cite{sofia2013mechanical} & Accelerometer array & Yes & Contact & Acceleration vector & Low spatial resolution\\ &&&&& Limited coverage \\
		\hline
		Shao et al.~\cite{shao2016spatial} & Accelerometer array & Yes & Contact & Acceleration vector & Tethered \\ &&&&& Does not admit large motion\\
		\hline
		Tanaka et al.~\cite{tanaka2015wearable,tanaka2017practical} & PVDF film & Yes & Contact & Acceleration scalar & Very low spatial resolution\\ &&&&& Physical units  unclear \\
		\hline
		Harrison et al.~\cite{harrison2010skinput} & Piezoelectric cantilever array & Yes & Contact & Acceleration scalar & No spatial coverage\\ &&&&& Low frequency bandwidth \\
	\end{tabular}
\end{table*}

The multimodal nature of touch sensing in the hand presents  challenges that have made it difficult to fully understand mechanisms of tactile sensing. It also provides inspiration for the design of biologically-informed artificial tactile sensors that has not been fully exploited to date.  There are more than a half dozen types of sensory receptors of touch in humans and other mammals  \cite{mcglone2014discriminative},  capturing different aspects and components of the mechanical signals supporting touch sensing and interaction \cite{visell2008tactile}. Sub populations of these receptors are most responsive to sustained mechanical stimulation (slowly adapting types, SA), and others to transient or vibratory signals (fast adapting, FA).  Receptors can be further categorized as sensitive to stimulation near to a mechancal contact   (types SA I and FA I),  or to more remote mechanical signals (SA II or FA II). Essentially all perceptual and motor functions of the human hand are enabled by input from several of these tactile submodalities, yielding a variety of information about skin-object contact~\cite{saal2014touch,jorntell2014segregation}.

While conventional accounts of biological touch sensing associate tactile perception with sensory resources near to the area of contact with an object (since these are the tissues that undergo the largest contact-induced deformations), recent research, including work in our own lab, reveals that mechanical contact with the skin elicits elastic waves that propagate throughout the hand in the tactile frequency range, from 0 to 800 Hz.  
These mechanical signals reach widespread vibration sensitive receptors (including type FAII receptors, also called Pacinian Corpuscles, PC), where they are transduced into neural signals that can encode contact forces, events, or  surfaces with which they originate \cite{delhaye2012texture,manfredi2012effect,shao2016spatial}.  This remote transmission of touch-elicited mechanical signals in the hand is very efficient at frequency ranges (approximately 20 to 800 Hz) that are relevant to vibration perception \cite{shao2016spatial,delhaye2012texture}. 

Human abilities of remote tactile sensing are especially associated with type FAII receptors (Pacinian Corpuscles, PC). PCs number in the many hundreds in each hand \cite{kandel2000principles}, are distributed throughout the limb, and are involved in tactile functions including texture discrimination, tool use, and the detection of object contact or slip.  However, it has been difficult to characterize the mechanical processes involved in human remote tactile sensing due to the complex array of tissues \cite{khatyr2004model} and physical regimes involved. 

Few electronic systems have been designed to capture  the array of distributed touch-related mechanical signals in the whole hand during natural tactile exploration, grasping, and manipulation in everyday activities.  As a result, we currently have limited understanding of tactile signals that are felt during such interactions. 
Knowledge of this type could help to elucidate the scientific basis of human touch sensing, as mediated by continuum mechanics of the hand, including effects of sensory loss accompanying disease.  It could also lead to advances in sensor engineering for robotics, upper-limb prosthetics, and other areas. An improved understanding of tactile signals gathered by the human hand is also needed in order to guide the design of tactile experiences elicited by emerging products and devices.

\subsection{Related Research}

Many tactile sensing devices are described in the literature, including several that have been applied to wearable and in-vivo tactile sensing (Table 1). However, the task of capturing transient mechanical signals in tissues of the entire hand presents difficult challenges due to  high density of tactile sensors, the distributed nature of the skin, the relatively large frequency range concerned,  the large dynamic range of displacements involved (spanning 5 orders of magnitude), and the kinematic complexity of hand movements that are involved in many hand activities.   
Techniques that have previously been investigated for capturing transient or vibratory signals in spatially distributed regions of soft tissues of the body include large area vibrometry methods, such as ultrasonic Doppler vibrometry \cite{sikdar2007ultrasonic,plett2001vivo}, laser scanning vibrometry \cite{rossi1995hand,xu2011vibration}, line laser sensors \cite{kawahara2006non},  high speed cameras \cite{tanaka2011noncontact,allen1999high}, among others. 
These typically require the hand to remain stationary, making it difficult to collect tactile data during hand activities, as we aim to accomplish.  

Many wearable sensing devices have also been previously investigated, including skin-wearable accelerometer arrays, such as the devices employed in our earlier work \cite{shao2016spatial,schafer2017transfer}, which are too kinematically constrained to admit most normal hand actions, partly motivating the device presented here. Others consist of sensor arrays with low spatial resolution, that have not been designed to transduce propagating signals throughout the hand  \cite{sofia2013mechanical,howe1989sensing,tarabini2012potential,verrillo1986effects,harrison2010skinput}. 
Another goal of prior research has been to design skin-like tactile sensors for strain \cite{wang2018skin,hua2018skin,yang2017three,do2017stretchable,cooper2017stretchable,dickey2017stretchable}, pressure  \cite{li2016assemblies,li2016mutual,li2016soft,zhu2018highly,nie2017high,park2014giant,gong2014wearable,pu2017ultrastretchable,trung2016flexible,chortos2016pursuing}, or similar mechanical signals. Due to the sensing principles used, these devices are unable to capture tactile signals with the large bandwidth required for capturing distributed vibrations in the hand.

 While the research presented here addresses human-wearable sensing, a related area of research involves the development of tactile sensors for robotics.  This problem has attracted considerable attention \cite{howe1993tactile,yousef2011tactile,nicholls1989survey,dahiya2010tactile}, and  might benefit from the integration of bioinspired, soft material substrates  that emulate the capacity of human skin to  channel the contact-produced mechanical signals of interest to distributed sensing locations.  The results presented here provide demonstrations, methods and data to guide the design of such devices.  From a design standpoint, however, even leaving aside our application to human sensing, the point of departure of our work from prior sensing efforts in robotics is our emphasis on the remote sensing of propagating vibrations, as exemplified by the human limb.

\subsection{Contents and Contributions of this Paper}

In this article, we present a new 126-channel wearable sensing instrument designed to investigate information content in mechanical signals propagating in the human hand, mirroring the capabilities of the network of vibration-sensitive mechanoreceptors that are widely distributed in hand tissues.  We designed the device to accommodate normal hand movements, a large frequency bandwidth, and spatial resolution sufficient to capture propagating mechanical waves in the regime relevant to touch perception.  The light weight and flexibility of the device allow it to be adapted to a variety of hand sizes and shapes, and to be used on either side of the hand (i.e., palmar or dorsal face).  We designed a custom \gls{fpga} based wearable \gls{daq} platform and communication protocol in order to interface with the accelerometer array and transfer the large-amount of resulting data to a computer in real-time.  
We demonstrate the ability of this device to capture wave propagation across the entire hand, and demonstrate methods for reconstructing these signals from measurements, yielding new insight into distributed information captured by the sense of touch, and a new model for tactile sensing via large area remote vibration sensing in the hand.   

The wearable sensing instrument presented here has many applications.  These include:
\begin{itemize}
\item Wearable systems for characterizing tactile signals felt by the hand during interaction with objects or haptic displays.  In this application, our sensing instrument may be used to quantitatively characterize tactile experiences for applications in product design, material selection, and haptic interface characterization.  The mechanical measurements may be integrated with software tool that is able to predict results of user testing.  Such a system may replace time-intensive product testing protocols or perceptual experiments.\\
\item Wearable devices for human-computer interaction that capture interactions between the hand and touched objects. In this application, the accelerometers in our device may  be used for both contact sensing and kinematic pose estimation (hand tracking).  Systems of this nature may be used in many applications for interactive computing and augmented and virtual reality. Results from the high resolution device presented here may also be used in order to optimize sensor selection and placement for a lower complexity and lower cost device suited for many practical applications.\\
\item Tactile sensing for robotic hands or other end effectors.  Such an application may also involve the integration of this device with a skin-like covering that provides mechanical coupling analogous to that in the human hand. In this application, our device may be used to endow a robot with the ability to capture rich tactile information via a distributed array of sensors that need not be positioned on the contacting surface of the robotic limb, improving the design of the robot.\\ 
\item Measurement systems for neuroscience studies.  Such an application would leverage the ability of our device to capture tactile signals from the whole, behaving hand, as in the results presented here.  The resulting data may be combined with software that precisely simulates the neural spiking responses of populations tactile neurons in the hand \cite{saal2017simulating}. This would enable new neuroscience paradigms connecting peripheral tactile signals to neural responses and representations and processing in the brain. 
\end{itemize}

\section{Tactile Sensor Array Design}
\label{sec:instrument_design} 

We designed a sensing instrument to capture the distributed  response of the skin  across the entire hand to multi-finger interactions of the naturally behaving hand, mirroring the vibration sensing capabilities of the biological hand.  Elasticity couples  skin motion at different locations, implying that it is only necessary to sample  at an array of  points with a nominal density determined by the spatial Nyquist frequency.  At the oscillation frequencies of interest (less than 1000 Hz), the wavelength is determined by the wave speed, which depends on the propagation regime (i.e., shear, surface, or compression) and tissue properties. Prior research suggests that tactile signals travel primarily as surface or shear elastic waves in soft tissues. Their wavelengths $\lambda$ are approximated by
\[
	\lambda = \frac{v_s}{f}, \ \ \ \ v_s = \frac{E}{2\rho (1+\mu)}
\]
where $v_s$ is the shear wave speed ($\approx$ 4.4-17.5 m/s,~\cite{manfredi2012effect}), $E$ is the elastic modulus ($\approx$ 0.13 MPa,~\cite{khatyr2004model}), $\rho$ is the density ($\approx$ 1.02 g/cm$^3$,~\cite{liang2010biomechanical}), and $\mu$ is Poisson's ratio ($\approx$ 0.5,~\cite{liang2010biomechanical}). 
In the regime relevant to tactile perception, $f < 1000$~Hz, and the wavelength $\lambda$ is greater than 1 cm. This motivates sampling the surface motion of the skin at a sparse array of points, as in the presented instrument.

\subsection{Electronics Design}

To achieve this, we designed the system to integrate  an array of 42 multi-axis accelerometers with wide frequency bandwidth. We capture data at high speed via an \gls{fpga}-based multichannel \gls{daq} board, with firmware on the FPGA and software on a PC, which may be a desktop or laptop system, or an embedded device. 
The accelerometer array is readily attached to the skin, capturing skin vibration (or other facets of vector acceleration, such as motion and gravity) over extended areas (Fig.~\ref{fig:AccIII_Position}). 
The \gls{daq} board receives accelerometer readings through 23 I$^2$C buses, and then transmits them to a PC or embedded system through a \gls{usb} cable. The measurement process is controlled via USB. 

In order to ensure wearability, the system was designed to be lightweight, with individual accelerometers having masses of 4.1 g. The flex PCB has a mass of less than 10 g. The \gls{fpga} board has a mass of 19.0 g, and dimensions ($87.0\times50.5$ mm) small enough to that it can be easily mounted on the arm by using a 3D printed, conforming bracket, as shown in Fig.~\ref{fig:VibrateFingerTip}(a).

\subsection{Accelerometer array}
The sensor array contains 42 three axis miniature accelerometers (Model LIS3DSH, ST Microelectronics). Each axis of the accelerometer has a selectable measurement range from  $\pm$2 to $\pm$16 g, with corresponding sensitivity ranging from 0.06 to 0.73 mg. The maximum output data rate is 1.6 kHz, ensuring that the bandwidth (up to 800 Hz) approximates the sensitivity range of the ensemble of low and high frequency sensitive mechanoreceptors in the hand. 
The small package of the accelerometer ensures a contact area less than $5.0\times5.0$ mm. Acceleration recordings are digitized and exported to the \gls{daq} board via I$^2$C buses. 

\begin{figure}[ht]
	\centering
	\subfloat[]{\includegraphics[width=43mm]{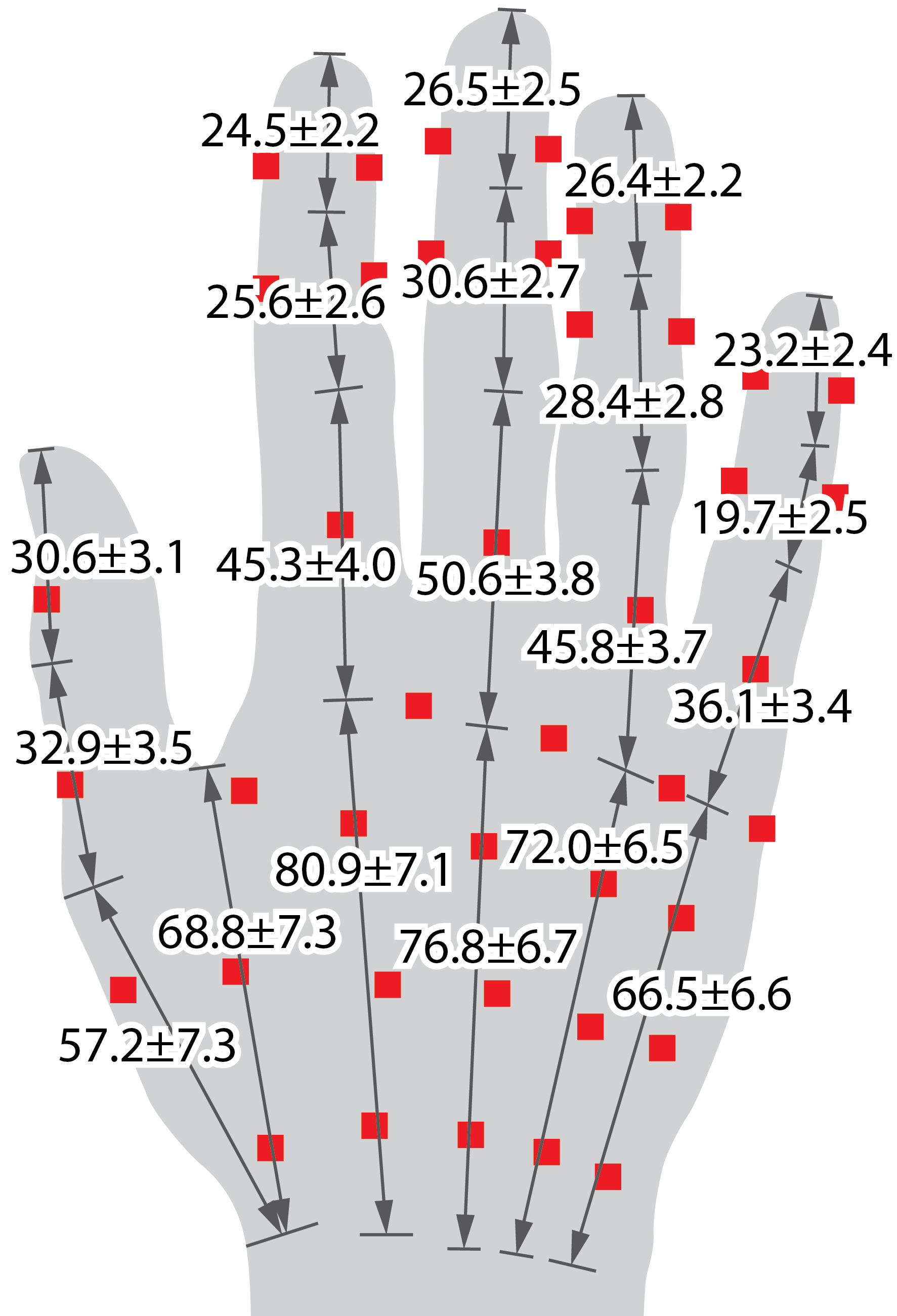}}
	\hfil
	\subfloat[]{\includegraphics[width=45mm]{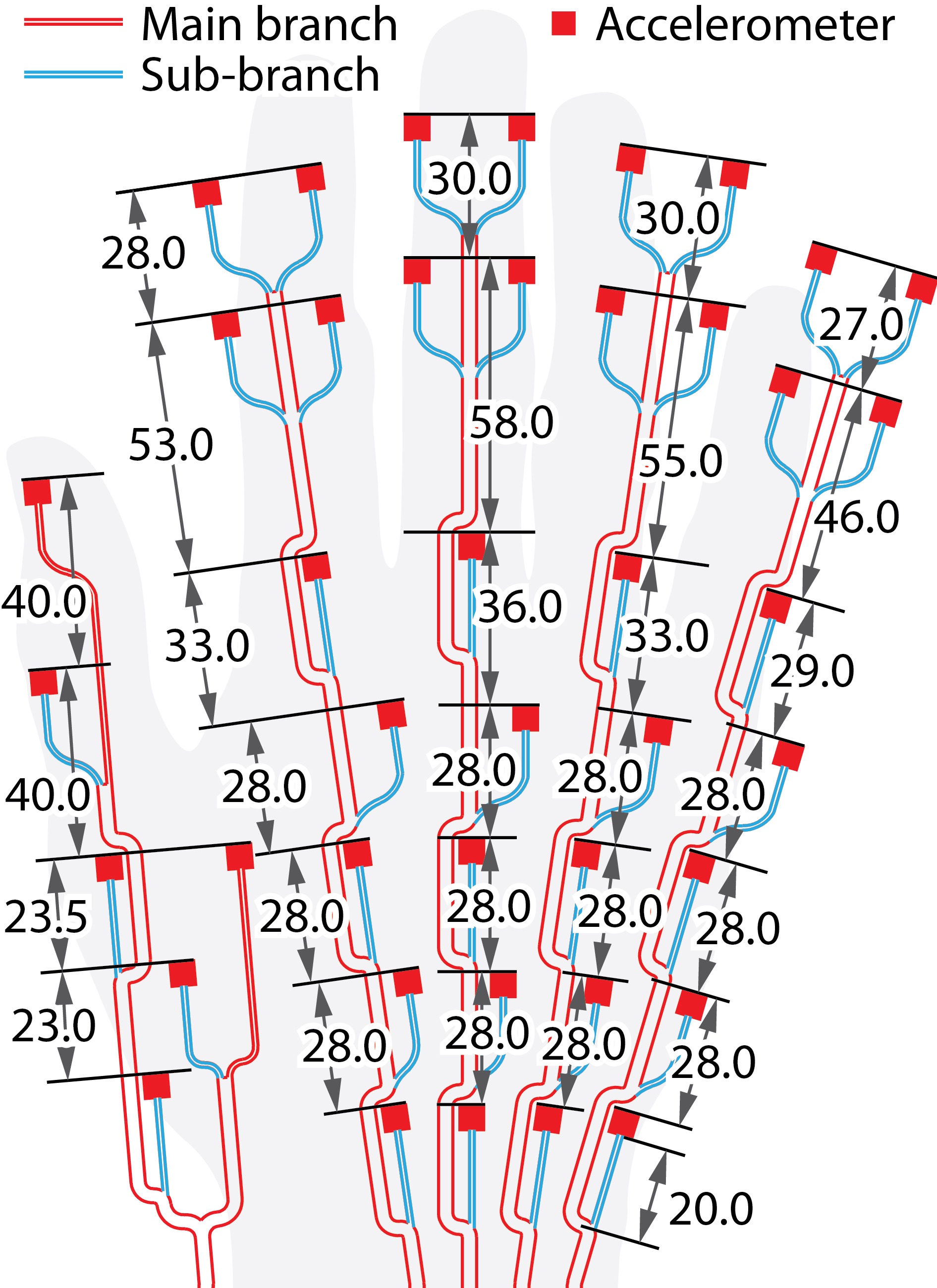}}	
	\caption{
		Design of the sensor array was based on hand anthropometry (mm). (a) Statistics (mean $\pm$ standard deviation) of dorsal hand anthropometric dimensions of 69 females and 70 males, with age ranging from 18 to 68 \cite{vergara2018dorsal}. Red squares indicated the example locations of the sensors mounted on the hand. (b) Designed dimension of the flexible PCBs. All sub-branch cables (blue) have length of 20 mm, in order to adapt the variance of hand sizes and admit unconstrained motion of the hand.
	}
	\label{fig:CableDesign}
\end{figure}

\subsubsection{Flexible PCB}
The accelerometer array and \gls{daq} system are connected via a set of five flexible \gls{pcb}s. The shape of the \gls{pcb}s is ergonomically designed with bowed regions that maximize flexibility, prevent cable contact interference during hand interactions, and that admit unconstrained motion in each of the segments of the hand. The same features ensured that the sensing array can be worn on hands of various sizes (Fig.~\ref{fig:CableDesign}(a)). The dimensions of the \gls{pcb}s (Fig.~\ref{fig:CableDesign}(b)) were determined based on studies of hand anthropometry \cite{vergara2018dorsal,imrhan2009hand}.
The sensors can be distributed over skin areas throughout the dorsal (back) surface of the hand (as indicated by the red squares in Fig.~\ref{fig:CableDesign}(a)), at locations where contact rarely occurs during manual interactions, but the device can also be mounted on the palmar surface should measurement conditions dictate. 
Standard temporary prosthetic glue is used to attach the accelerometers to the skin, to ensure a consistent flexible bond over a small contact patch. Thin prosthetic tape is a substitute for adhesives, enabling more rapid attachment. The accelerometer array can also be integrated into a glove to increase flexibility of putting on and removing the sensors. 

\begin{table}[hb]
	\renewcommand{\arraystretch}{1.3}
	\caption{Hand anatomical locations of the sensor.}
	\label{tab:Anatomy_Location}
	\centering
	\begin{tabular}{ll}
		Anatomical location & Accelerometer number\\
		\hline
		Distal phalanges (DP) & 1, 2, 11, 12, 21, 22, 31, 32, 45.\\
		Middle phalanges (MP) & 3, 4, 13, 14, 23, 24, 33, 34.\\
		Proximal phalanges (PP) & 6, 16, 26, 36, 46.\\
		Metacarpals head (MH) & 5, 15, 25, 35.\\
		Metacarpals shaft (MS) & 8, 18, 28, 38, 44.\\
		Metacarpals base (MB) & 7, 17, 27, 37.\\
		Carpals (C) & 9, 19, 29, 39, 43. \\
		Between MS  of digits I, II & 41.\\
		Between MS and MB of digit II  & 42.\\
		\hline
	\end{tabular}
\end{table}

The PCB layout is designed based on hand anatomy and motion considerations  (Tab.~\ref{tab:Anatomy_Location}), accounting for movements of all five digits and the back (dorsum) of the hand, including  the following anatomical regions: distal, intermediate and proximal phalanges, the metacarpal and carpal areas (Fig.~\ref{fig:AccIII_Position}). 
To avoid artifacts, we avoided positioning the sensors on joints, where the skin surface deforms most during digit movement.  Instead, sensors are  positioned on or between bones. This also reduced interference introduced by  movement, and ensured consistent attachment of the sensors. 
The device is divided into five independent PCBs, or main branches, each corresponding to one digit. 
The cable morphology was revised through several iterations using paper mockups so as to impose minimal constraint to finger movements. 

\begin{figure}[!t]
	\centering
	\includegraphics[width=83mm]{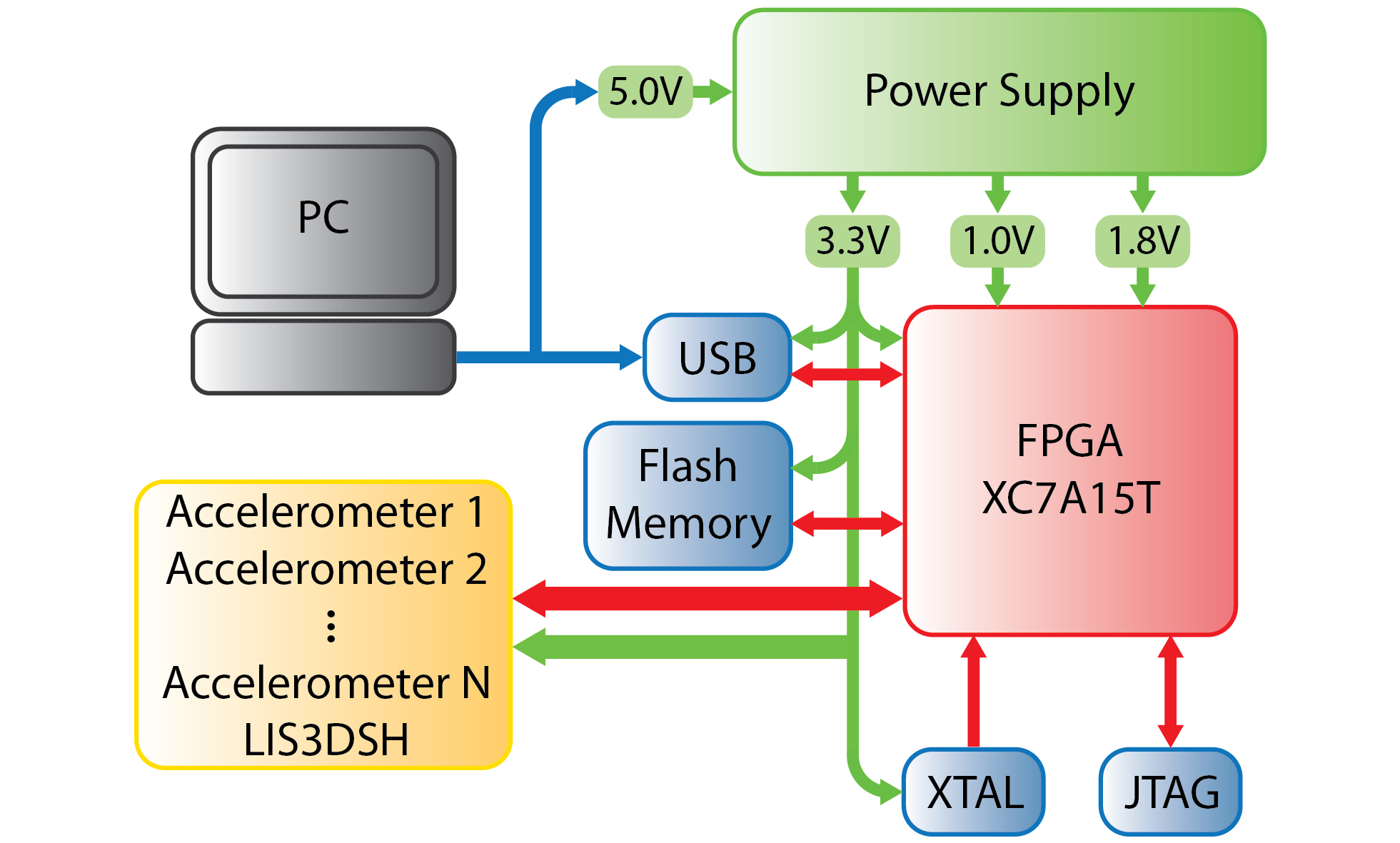}
	\caption{
		Overview of the wearable sensing system. N is the number of accelerometers that are connected to the FPGA through I$^2$C buses on flexible PCBs. $N\in$ \{6, 9, 15, 18, 24, 27, 33, 36, 42\}. The data is streamed in real time to a PC or embedded device via a USB connection, which also supplies the power for the instrument. 
		The green and red arrows indicate power connection and communication signal lines respectively. Power and communication are both handled over USB, as depicted as blue arrows.
	}
	\label{fig:System_diagram} 
\end{figure}

\begin{figure}[!t]
	\centering
	\includegraphics[width=83mm]{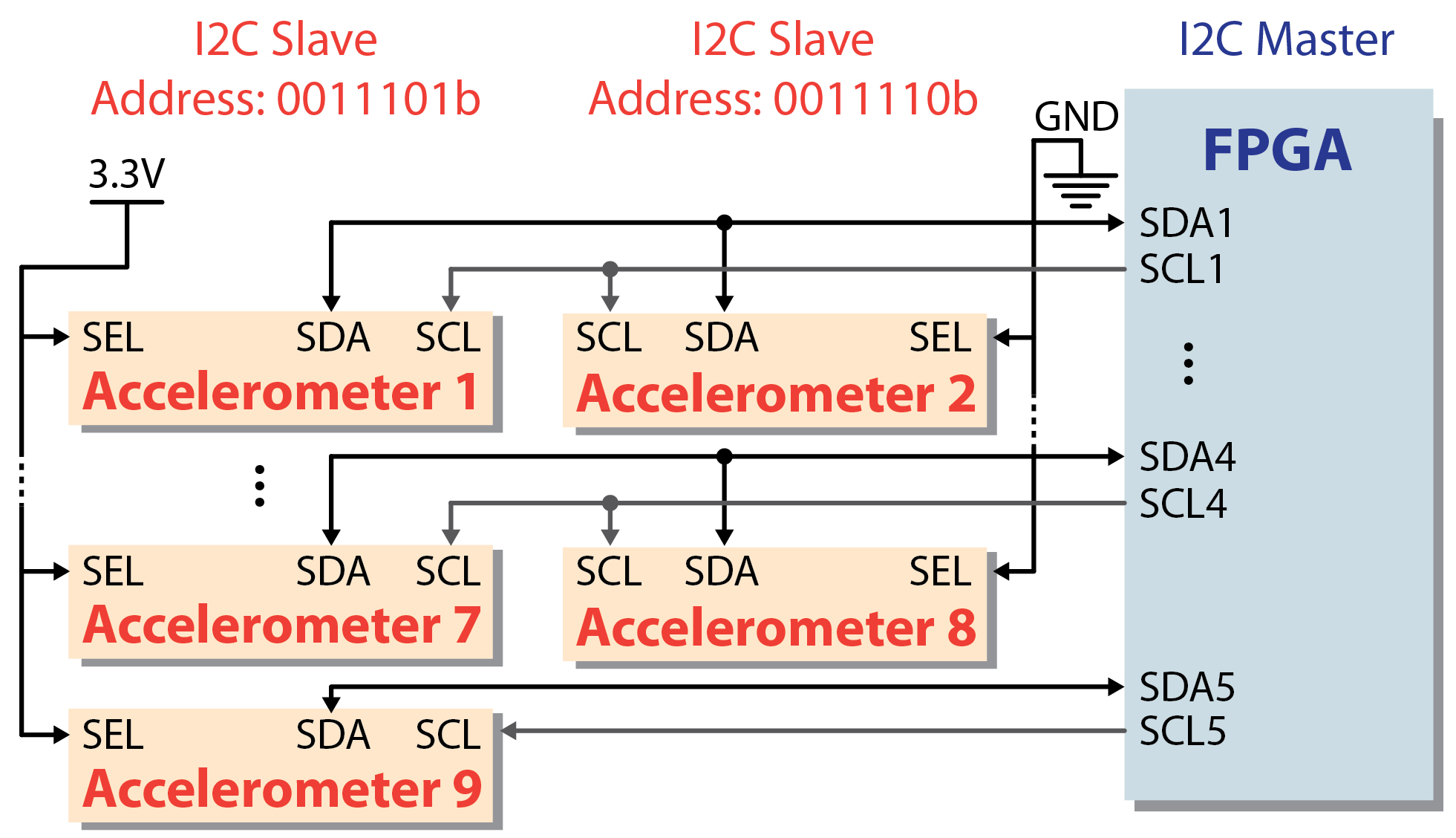}
	\caption{
		I$^2$C communication diagram between the FPGA (master) and accelerometers (slaves). Each I$^2$C bus is shared by one pair of accelerometers that communicate with the FPGA alternatively.
		SCL: Serial clock line. FPGA outputs a clock of 1.6 MHz (maximum) to each pair of sensors. 
		SDA: Serial data line.
		SEL: I$^2$C address selection. For each pair of accelerometers, the one with this pin pulled up to 3.3V has I$^2$C address of 0011101b, while the pin of the other one is grounded to have I$^2$C address of 0011110b.
		Power supply lines for accelerometers are not displayed in this diagram. 
		Only one branch of the flexible PCB, consisting of accelerometer 1 to 9, is shown in this diagram. 
	}
	\label{fig:I2C} 
\end{figure}

Each sensor was connected to a main branch cable (3.0 mm width, connection of all sensors) through a sub-branch  (1.3 mm width, connected to one sensor only). When the sensors are affixed to the hand, the sub-branch cables lift the branch cable above the hand to prevent contact with the skin, allowing  for flexion and extension of digits. Those cables were designed to have a curved shape to have better flexibility and facilitate \gls{pcb} layout.
Unlike the \gls{dp} and the \gls{mp} where the sensors were placed on the side, The \gls{pp} had a sensor on top due to space limitations on their side in most normal hand interactions.

\begin{figure*}[!t]
	\centering
	\includegraphics[width=175mm]{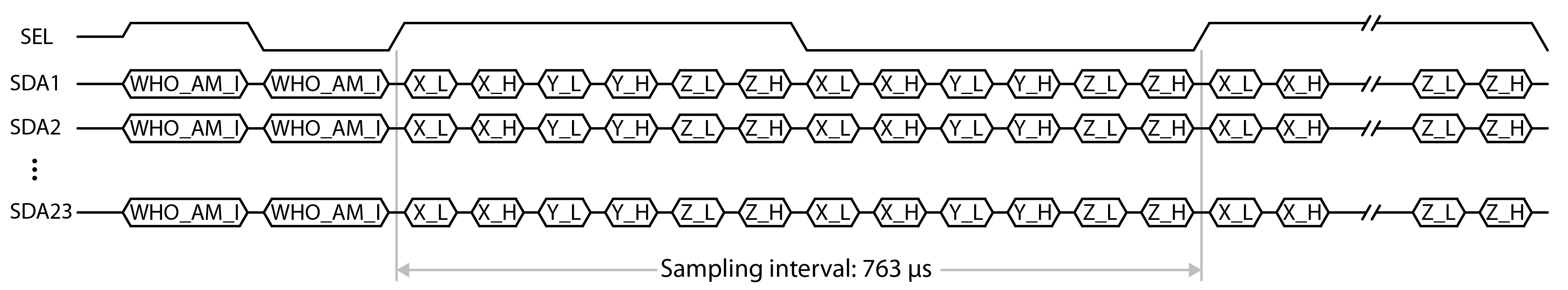}
	\caption{
		Data sampling protocol of the wearable sensor array through 23 I$^2$C buses. WHO\_AM\_I register of each accelerometer is checked before each measurement. Then, data is sampled from accelerometer SEL pull-up and pull-down groups alternatively. For a sample of an accelerometer, X,Y and Z axis data is acquired sequentially, each with the low byte (L) followed by the high byte (H).
	}
	\label{fig:I2C_Sequence} 
\end{figure*}

The sensors are divided into five branches, each connected to a dedicated port on the DAQ board. All of them operate independently. The system remains functional if any of the five branches is removed, as can be accomplished by trimming the connective area of the \gls{pcb}. This feature was designed to accommodate demands for task-specific or localized measurement, for example, during interaction of a single digit. 
 Each branch has independent circuit connection with the \gls{fpga} board, with five independent I$^2$C buses, with power and ground connections. All five branches are physically connected at the end of the PCB, near the connectors, and can be separated easily by cutting through a non-functional area between branch circuits (Fig.~\ref{fig:AccIII_Position}). This makes it possible to deploy the device in configurations that need not include all fingers, or involve other usages not presented here, such as simultaneous sensing on both faces of selected digits, on other regions of the limb or body.

The PCB design accommodates sensor configurations that  vary depending on the size and shape of the hand on which they are mounted, with locations on the hand surface and relative orientations that are registered to the anatomical features of the hand.  For any sensor placement and fixed posture of the hand, the configuration of each of the $N_s$ sensors are described by a set of spatial positions and  orientations, in three dimensions, which can be expressed as pose matrices $T_i$ given by 
\begin{eqnarray}
T_i &=& \begin{pmatrix}
  R_i & \pp_i \\
  0_{1\times 3} & 1  
\end{pmatrix}
\end{eqnarray}
with
\begin{eqnarray}
 \pp_i &=& (p_{i,x}, \ p_{i,y}, \ p_{i,z})^\top, \\
 R_i &\in& SO(3), \\
 i &=& 1, 2, \ldots, N_s 
\end{eqnarray}
where $\pp_i$ are the coordinates of the $i$th sensor relative to a fixed coordinate system of the spatial environment,  and $R_i$ is a $3\times 3$ rotation matrix,  which specifies  an orthonormal frame, describing the orientation of the $i$th sensor relative to this environment. Both $\pp_i$ and $R_i$ depend on the kinematic pose of the hand, and the relative position and orientation of different sensors can change depending on the hand position orientation, and the pose of the fingers.  
In section III below, we describe how to  map the sensor configuration onto a standard reference hand, for the purpose of  integrating and analyzing tactile signals across the geometric surface of the hand.

\subsection{FPGA-based multichannel data acquision board} 

We designed an \gls{fpga} board (model XC7A75T, Artix-7 series, Xilinx inc.) to acquire data from the 42 accelerometers. An overview of the system is shown in Figure~\ref{fig:System_diagram}.
 In order to capture data from the ensemble of sensors in real time, the FPGA was programmed to interface with the accelerometers through 23 I$^2$C bus connections. Those buses are used to transmit sensor recordings to the \gls{fpga} in a parallel manner. They are divided into five branches, corresponding to sensor mounting locations on the five digits. The branches on digit II to V each include 5 I$^2$C buses, while the branch on digit I contains 3 buses. 
The bus branch on digit V contains accelerometers 1 to 9, with each I$^2$C bus communicating with two accelerometers except the last one, which is connected to accelerometer 9 only (Fig.~\ref{fig:I2C}).
Similarly, branches on digit IV, III and II contain accelerometer 11 to 19, 21 to 29 and 31 to 39 separately. The number 10, 20, 30 and 40 were skipped as they have no I$^2$C connection. The branch on digit I contains three I$^2$C buses, connecting accelerometer 41 to 46. 
All branches have isolated circuit routes and interface with the \gls{fpga} via separate AXT6 pitch connectors on the board, which enables separation of any branch from the others. 
The board includes a JTAG module (IEEE 1149.7) that makes it possible to interface the board with a PC via JTAG-SMT2, for programming and debugging purposes.
A N25Q256A flash memory module enables the \gls{daq} program to boot and operate without JTAG connection with the PC. 

Our firmware design enables data to be streamed to the PC in real time through a \gls{usb} cable, which also supplies power to the \gls{fpga} board. We have interfaced the device with a desktop PC and with embedded computers running Windows and Linux operating systems. 

\subsection{Communication protocol and data storage} 
Communications between the accelerometers and \gls{fpga} follow a data sampling protocol shown in Figure~\ref{fig:I2C_Sequence}.
In order to maximize the number of sensors that can be interfaced, we index sensors on each I$^2$C bus via the SEL bit.  For any pair of accelerometers connected to the same I$^2$C bus, the data are  in alternating fashion. Each is  associated with a different slave address indexed via the SEL bit. Accelerometers with SEL  high (pin voltage 3.3V) are associated with an I$^2$C address with SEL = 1, while those with SEL pin grounded have an address with SEL = 0. 
Data are acquired in interleaved fashion, first from the 23 odd numbered accelerometers (SEL = 1), then from the remaining 19 even numbered accelerometers (SEL = 0).

One sample of data from an accelerometer returns data $(a_x, a_y, a_z)$ corresponding to the cartesian components of acceleration in the local frame given by the orientation of the accelerometer (as we demonstrate below, this data can be interpreted for analysis even if the orientation frame is unknown). Before reading a sample, data availability of all accelerometers is verified via the status registers, which are designed to synchronize data transmission among the I$^2$C communication buses. Although the latter have independent clock lines, these lines are all synchronized and clocked to 1.6 MHz. After accounting for communication overhead, the sampling frequency is 1310 Hz.  Data from all sensors in the array (including both odd and even indexed sensors) is captured within one period at this rate. Accelerometer data received by the \gls{fpga} are transfered to the PC via \gls{usb} and stored in a binary file in real time. 

\section{Validation Experiments and Results}\label{sec:test_and_result} 
We investigated the performance of the sensing system by examining data captured by the sensors, and by investigating  the  applicability of this device to capturing distributed vibrations in the skin, including vibration signatures produced through touch contact.  We assessed the  device in both bench-top experiments that directly probed the performance of the sensors in the array, and through testing as the device was worn on an interacting human hand, while the latter performed a variety of gestures.

\begin{figure}[!t]
	\centering
	\subfloat[]{\includegraphics[width=83mm]{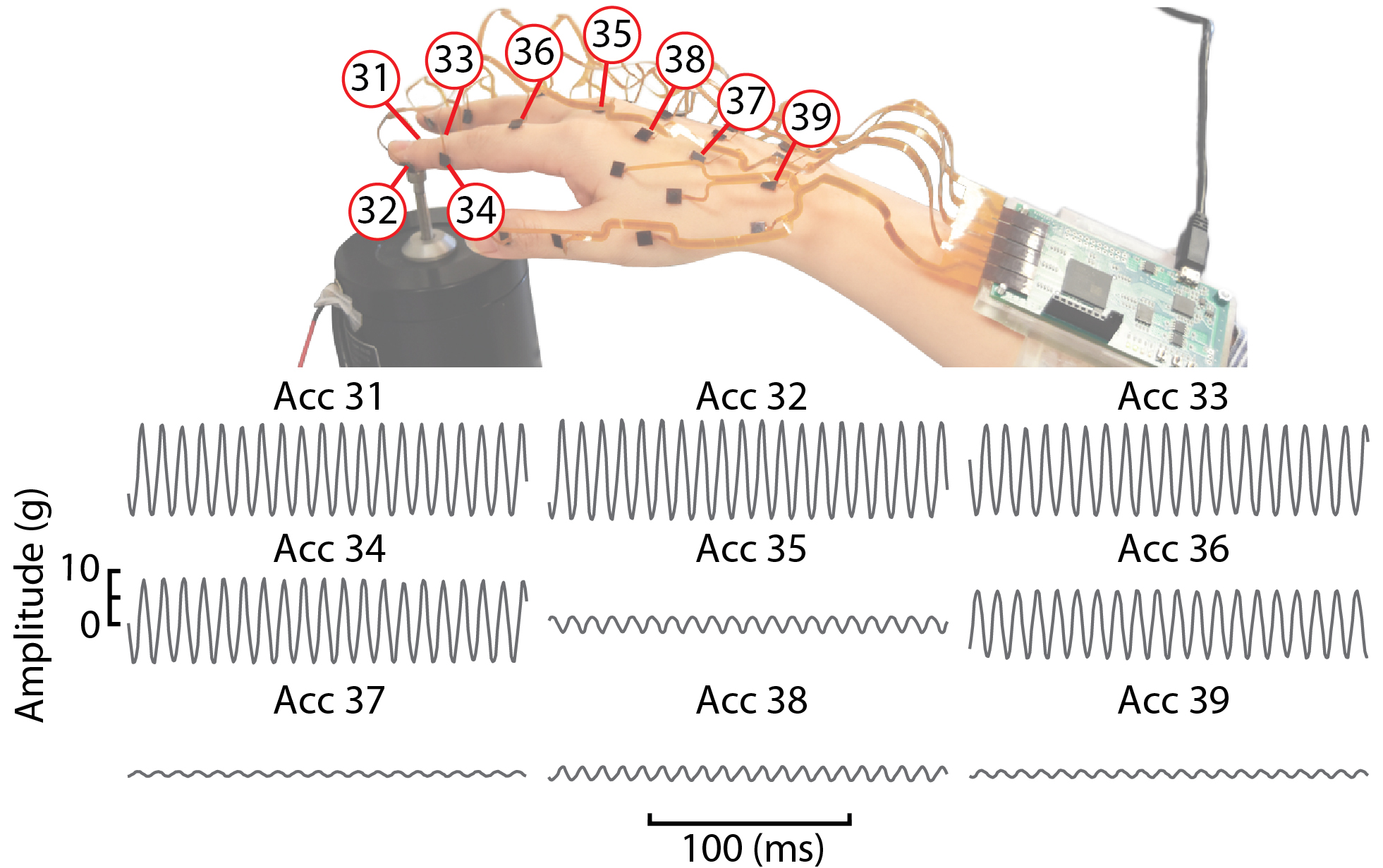}}
	\hfil
	\subfloat[]{\includegraphics[width=80mm]{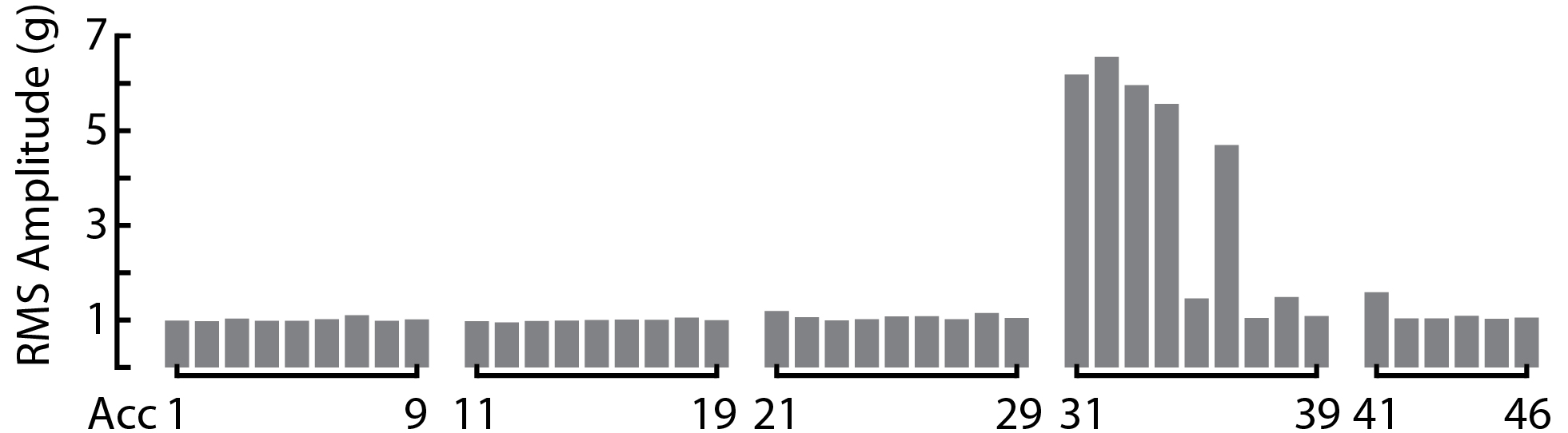}}
	\hfil
	\subfloat[]{\includegraphics[width=80mm]{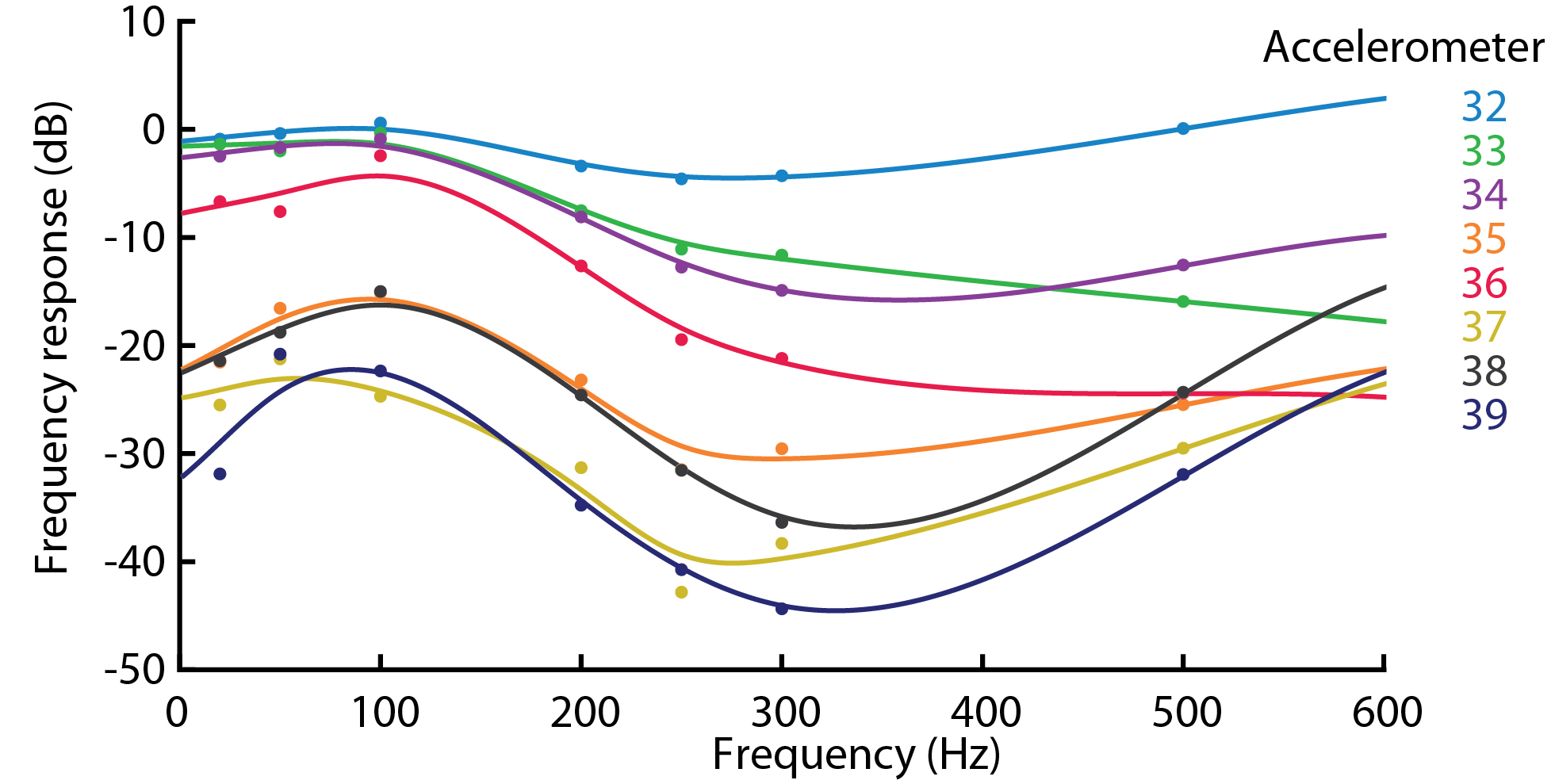}}
	\hfil
	\caption{Skin frequency response at sensor locations. (a) A 100 Hz sinusoidal stimulus was applied to the tip of digit II of the subject. Acceleration in the direction with the highest vibration amplitude $s_i(t)$ is shown for accelerometer 31 to 39 (digit II branch). (b) \gls{rms} amplitude measured at all locations. (c) Location-dependent frequency response of accelerometer 32 to 39 with respect to accelerometer 31.}
	\label{fig:VibrateFingerTip}
\end{figure}

\subsection{Sensor Array Performance} 


The accelerometers produce vector signals $\mathbf{a}_i(t)=(a_{i,x},a_{i,y},a_{i,z})$, where $i$ indexes the sensor and $t=0,\tau,2\tau,\ldots, N\tau$ indexes time.  The sample period $\tau = 1/f_s = 0.77$ milliseconds.  In order to integrate information from the ensemble of these vector measurements for the analysis, the positions $\pp_i$ and orientations $R_i$ of each sensor should be known.  While the former is known relative to anatomical landmarks, the latter is more challenging to acquire, as the orientation depends on the pose of the hand and orientation of the skin.  To address this, we developed a Principal Components Analysis based method for computing an orientation-invariant scalar value that captures the most salient available at each sensor.  The method is to project the vector sensor signal onto an estimate of its instantaneous principal component. This operation is linear, avoiding nonlinear artifacts that would be introduced by computing a magnitude, and preserves crucial phase information.  Using data $X_i$ captured from each sensor over a sliding window of hundreds of samples, for each sensor, we first compute the  
  eigenvector $\mathbf{w}$ of the  covariance matrix $\mathbf{X}_i = \{\mathbf{a}_i(0) \ \mathbf{a}_i(\tau) \ \cdots \ \mathbf{a}_i(N\tau) \}$ with the largest eigenvalue $\lambda$, where $N$ is the number of measurements (typically, $100<N<1000$, corresponding to a small fraction of a second).   By abuse of notation, we refer to this scalar value as $a_i(t)$, where 
\begin{equation}
a_i(t) = \mathbf{a}_i^T \mathbf{w}, \ \ \mathbf{w} =  \arg \max \left\{ \frac{\mathbf{w}^T \mathbf{X}^T \mathbf{X} \mathbf{w}}{\mathbf{w}^T\mathbf{w}} \right\}
\label{eq:waveprojection}
\end{equation} 
No temporal smoothing or filtering is performed.  With this approach, the projected scalar acceleration value computed from each sensor signal depends on the known position, but not the unknown orientation, of the sensor.  It also depends on the applied stimulus, which can differentially excite vector skin motion at different locations.

\begin{figure}[!b]
	\centering
	\includegraphics[width=73mm]{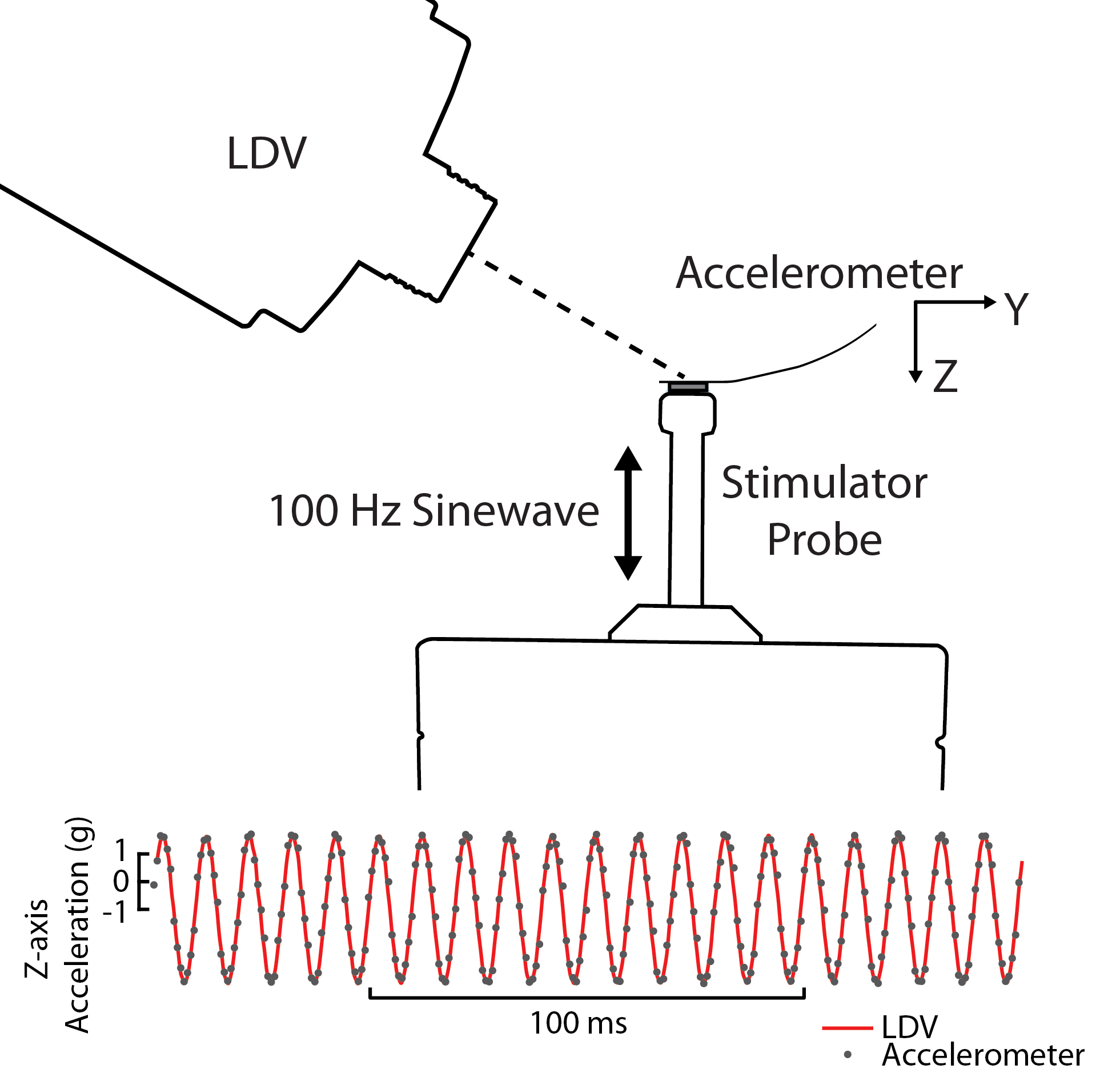}
	\caption{
		A signal waveform of one accelerometer mounted on the tip of an actuator, with Z-axis pointing along the actuator's vibration direction. The actuator output is a 100Hz sinusoid. Comparison with LDV acceleration measurement along Z-axis direction was made.
	}
	\label{fig:Acc_vs_LDV} 
\end{figure}

We tested the device performance when worn on a human hand (Fig.~\ref{fig:VibrateFingerTip}) as the hand was stimulated by a 100 Hz sinusoidal stimulus applied to the tip of digit II of the subject using a controlled electromechanical stimulator (Model 4810, Br\"{u}el and Kj{\ae}r). The hand was supported with digit II extending to contact the steel probe of the stimulator. The output of the ensemble of sensors was captured, and scalar acceleration signals computed for each accelerometer.  
A 200 ms segment of the values captured from one branch of the accelerometer array, representing one-fifth of the accelerometers, is shown in Figure~\ref{fig:VibrateFingerTip}. The branch corresponds to digit II, in the configuration shown in panel (a) of the figure. The signals measured at distributed points in the skin retain a highly sinusoidal waveform, reflecting the fact that signal propagation in hand tissues is highly linear at these frequencies.  The amplitude decreased with increasing distance, reflecting energetic losses due to damping in hand tissues.


We also validated the measurements captured with individual sensors in our device using an accurate, non-contact laser doppler vibrometer (Model PDV-100, Polytec).  
The \gls{ldv} had a sampling frequency of 22 kHz. The measurement range was set to be 100 mm/s.  The maximum acceleration was 13800 m/s$^2$, and the resolution was 0.02 $\frac{\mu\text{m}}{\text{s}}/\sqrt{\text{Hz}}$. 
A single accelerometer of the device was attached to the probe of the exciter vibrating (100 Hz sinusoidal signal) along Z-axis of the accelerometer. Thin reflective tape was attached to the top of the accelerometer board, where the LDV was directed.
Readings of all accelerometers were recorded through the device.  We compared the output of the sensor that was attached to the probe with the signal from the  \gls{ldv}.
The recording of the \gls{ldv} was low-pass filtered at 1000 Hz to better match the bandwidth of the sensor array, and differentiated to obtain  acceleration, which was compared with the $a_z(t)$, the accelerometer measurement component along the Z-axis  (Fig.~\ref{fig:Acc_vs_LDV}). The results indicate that measurements of the accelerometer and \gls{ldv} were in close agreement. 

\begin{figure}[!t]
	\centering
	\includegraphics[width=78mm]{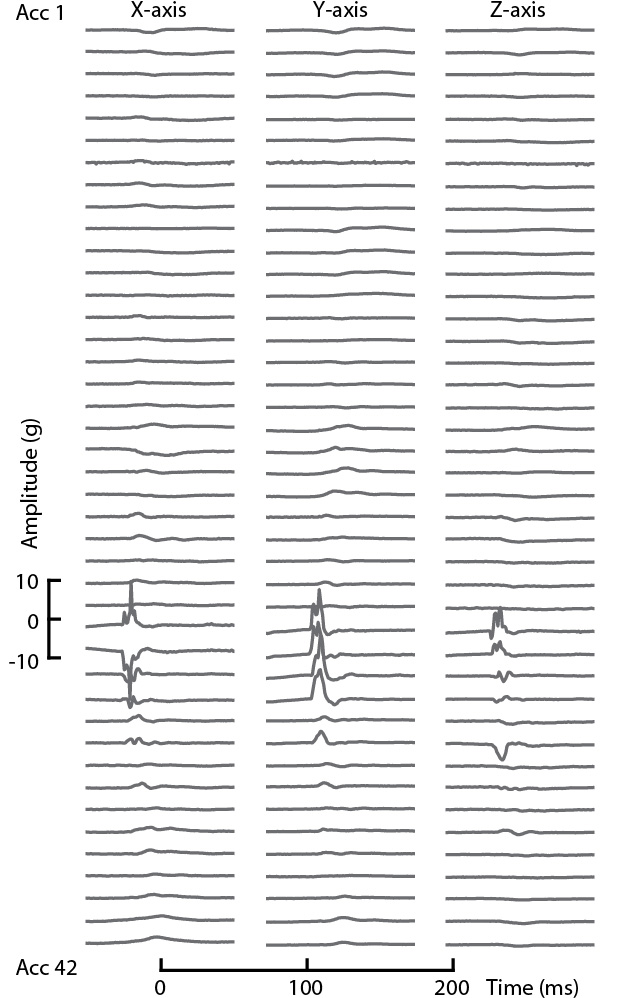} 
	\caption{X, Y and Z axis signal waveforms for all 42 accelerometers (one per row, numbered from top to bottom) during tapping digit II. }
	\label{fig:TapII} 
\end{figure}

\subsection{Wearable Sensing Experiments} 

We further evaluated the ability of the system to accommodate natural movements and touch contacts of the hand using a representative variety of hand actions, for the purpose of capturing distributed vibrations (Fig.~\ref{fig:TapII}) excited by skin-object contact during everyday interactions (Fig.~\ref{fig:AccIII_Demo}).  Here, we emphasize the ergonomic aspects of the device.  This shows that with the sensor array configured on the hand as shown, a wide variety of manual interactions is possible,  without constraints on the motion of the hand or digits. 
\begin{figure}[!h]
	\centering
	\includegraphics[width=73mm]{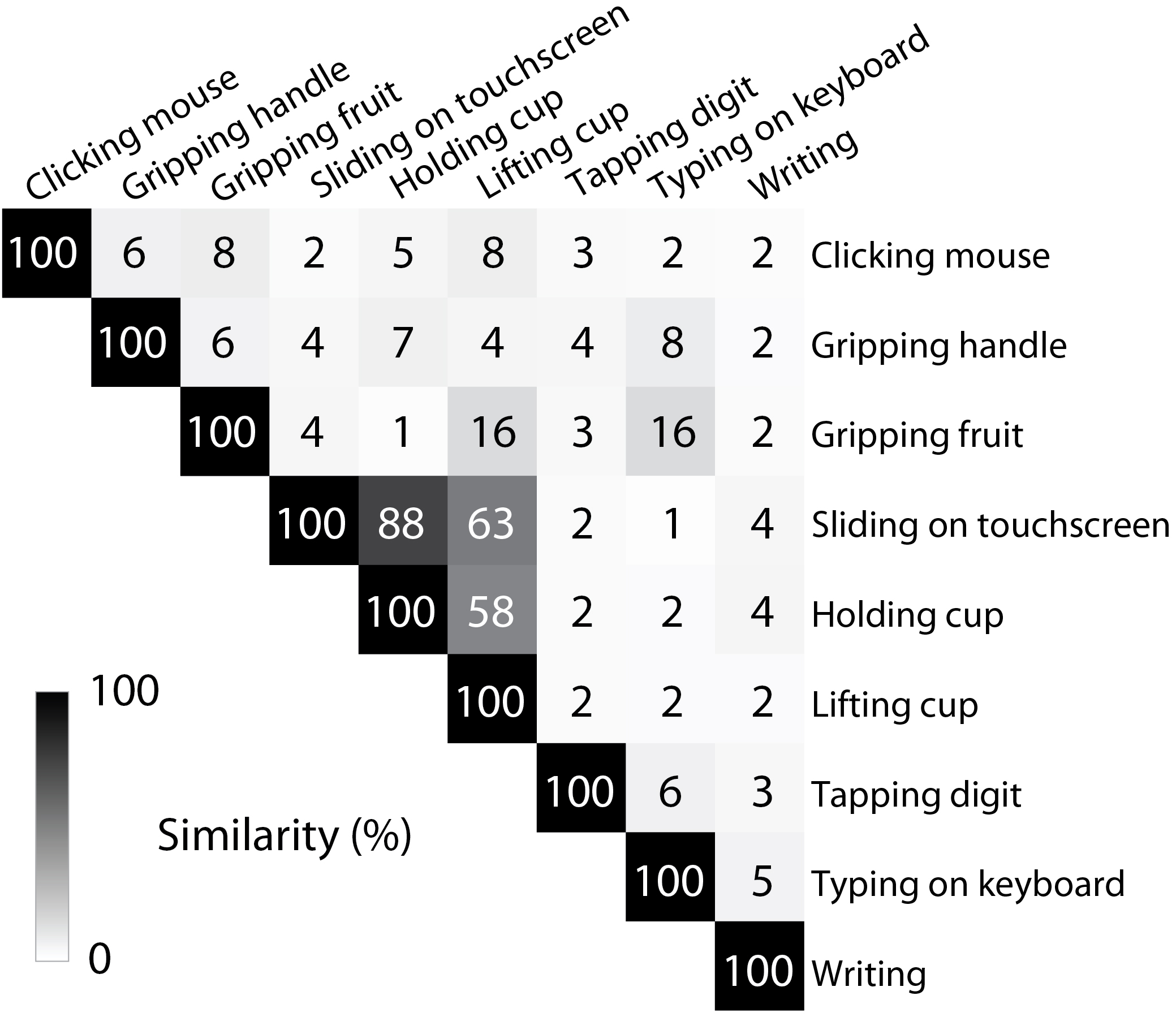}
	\caption{Similarity computed from sum of maximum correlation between tactile signals elicited by different gestures.}
	\label{fig:SimilarityMatrix} 
\end{figure}
In addition, we evaluated the similarity between tactile signals elicited by different gestures (Fig.~\ref{fig:SimilarityMatrix}), based on measurements of $N_s=42$ accelerometers. To compare the  signals elicited by one gesture $A(t) = \{a_1(t) \ a_2(t) \ \cdots \ a_{N_s}(t)\}$ to the signals elicited by another gesture $\bar{A}(t) = \{\bar{a}_1(t) \ \bar{a}_2(t) \ \cdots \ \bar{a}_{N_s}(t)\}$, we computed a similarity score given by
\begin{equation}
S(A,\bar{A}) = \frac{\sum_{i=1}^{N_s} \max_{\tau}(|\int_{t}a_i(t)\bar{a}_i(t+\tau)dt|) /\sigma_{a_i}\sigma_{\bar{a}_i} }{\sum_{i=1}^{N_s} \int_{t}\bar{a}_i(t)^2dt / \sigma_{\bar{a}_i}^2}
\label{eq:similarity}
\end{equation}
where $\sigma_x$ is the standard deviation of a signal $x$. The results indicates how tactile signals elicited by different actions are highly distinct.  The tactile signals that were most similar (e.g., grasping a cup or sliding fingers on a handheld touch screen) were those that engaged similar combinations of digits in similar configurations.


\begin{figure*}[ht]
	\centering
	\subfloat[]{\includegraphics[width=40mm]{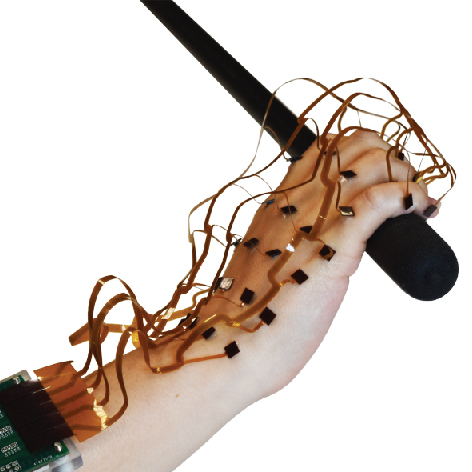}}
	\hfil
	\subfloat[]{\includegraphics[width=40mm]{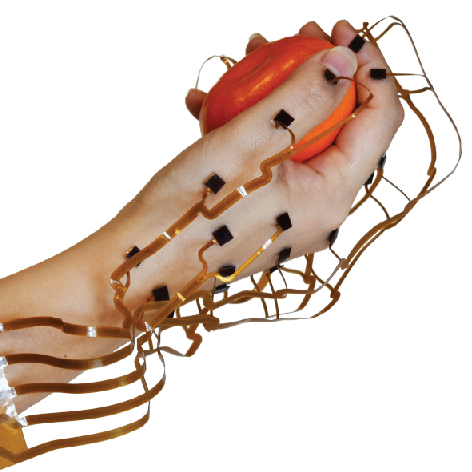}}
	\hfil
	\subfloat[]{\includegraphics[width=40mm]{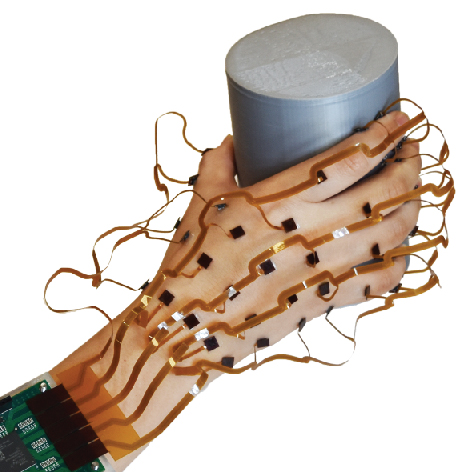}}
	\hfil
	\subfloat[]{\includegraphics[width=40mm]{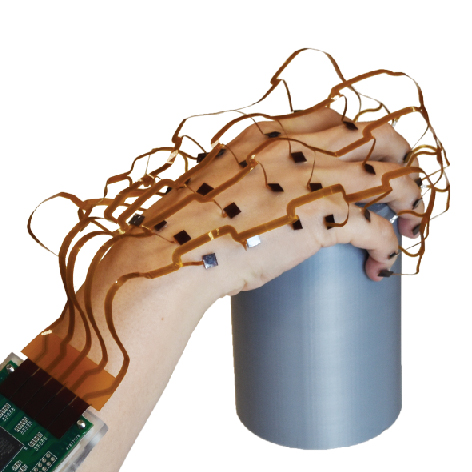}}
	\hfil
	\subfloat[]{\includegraphics[width=40mm]{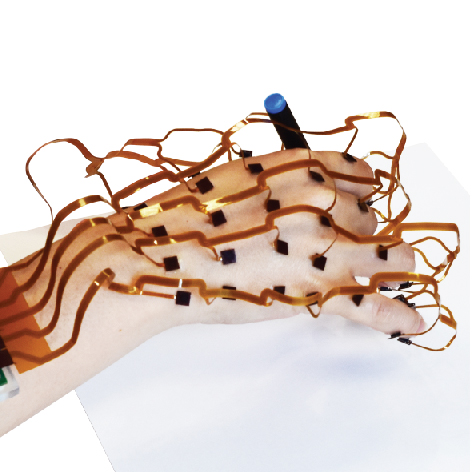}}
	\hfil
	\subfloat[]{\includegraphics[width=40mm]{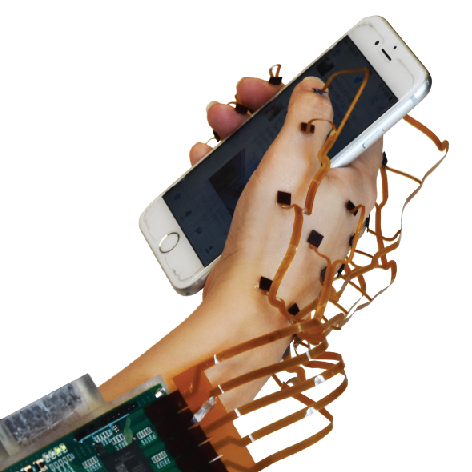}}
	\hfil
	\subfloat[]{\includegraphics[width=40mm]{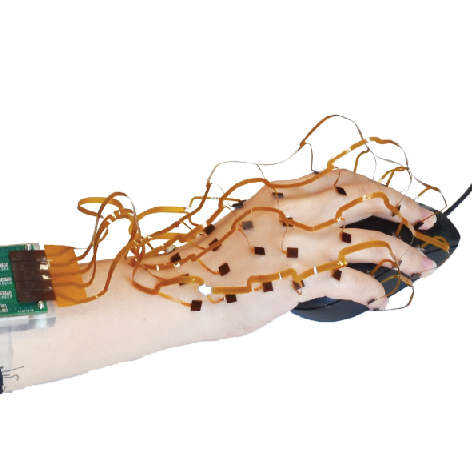}}
	\hfil
	\subfloat[]{\includegraphics[width=40mm]{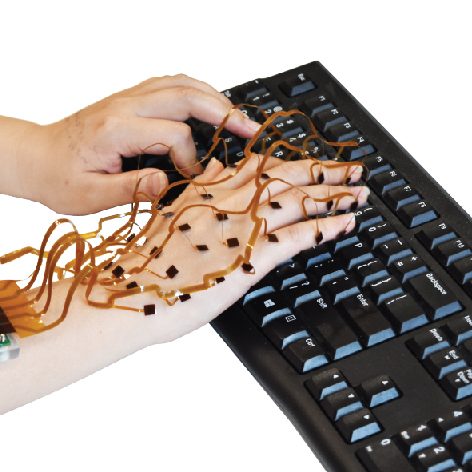}}
	\caption{ 
		Capturing tactile signals when performing natural hand gestures. There were no contact between cables and the skin. (a) Gripping handle. (b) Gripping fruit. (c) Holding cup. (d) Lifting cup. (e) Writing. (f) Sliding on touchscreen. (g) Clicking mouse. (h) Typing on keyboard. 
	}
	\label{fig:AccIII_Demo}
\end{figure*}

\subsection{Reconstructing Surface Wave Propagation in the Hand} 
\begin{figure*}[ht]
	\centering
	\subfloat[]{\includegraphics[width=34mm]{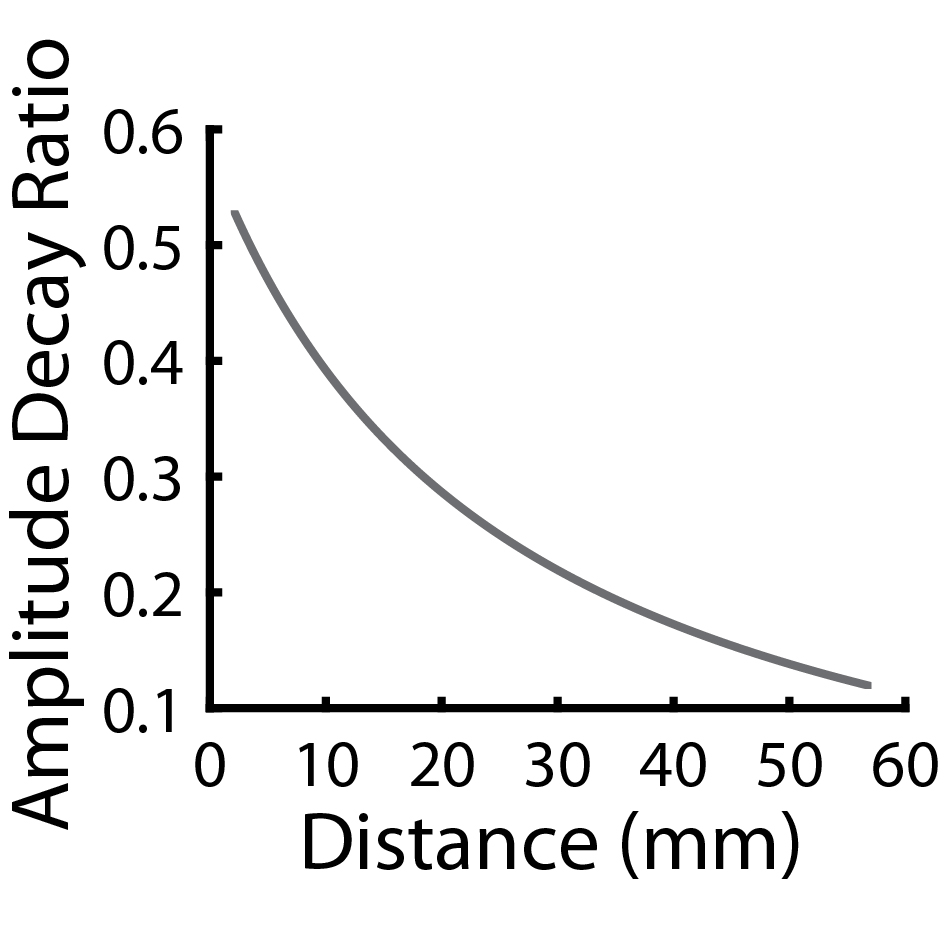}}
	\hfil
	\subfloat[]{\includegraphics[width=30mm]{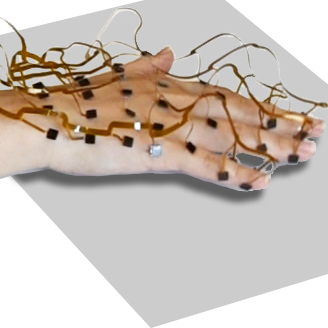}}
	\hfil
	\subfloat[]{\includegraphics[width=30mm]{fig/AccIII_Demo_sub-01}}
	\hfil
	\subfloat[]{\includegraphics[width=170mm]{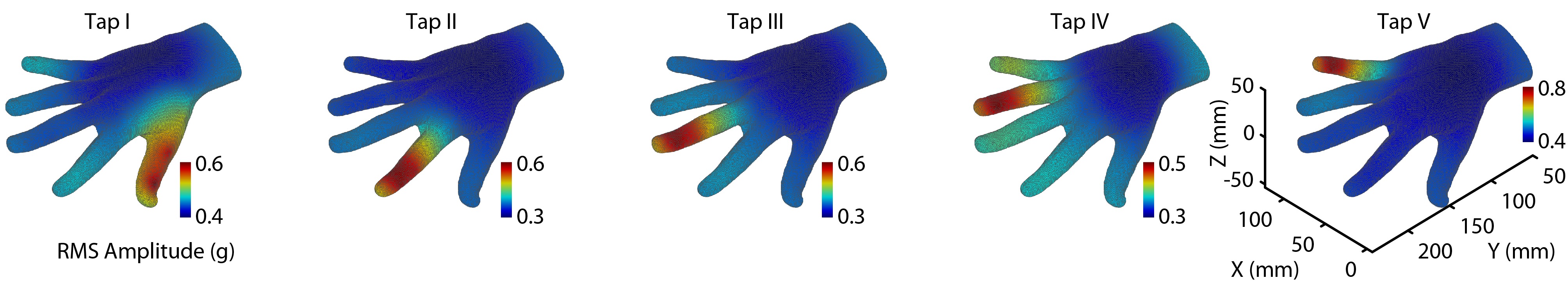}}
	\hfil
	\subfloat[]{\includegraphics[width=170mm]{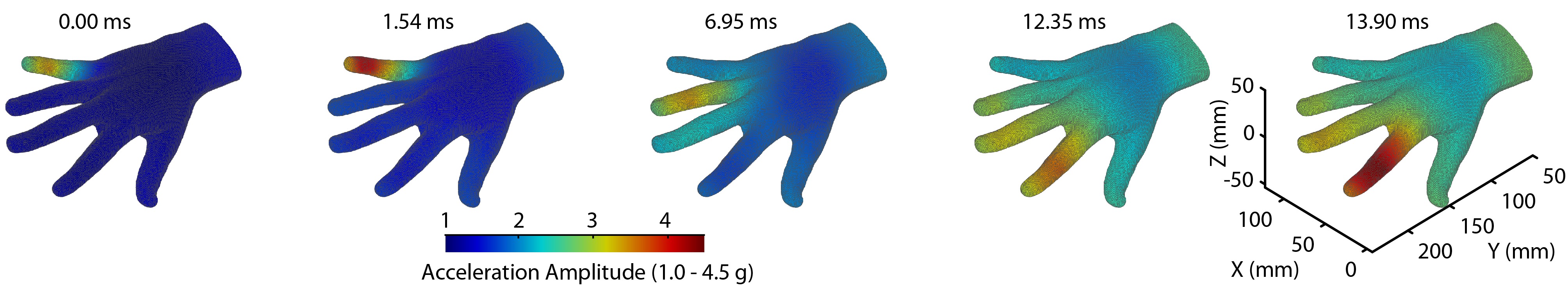}}
	\hfil
	\caption{(a) From physiological data in the literature \cite{manfredi2012effect}, we fit the local relationship between the  amplitude of propagating tactile signals, or waves, and their propagation distance. As described in the text, we used this information in order to interpolate between measurements at different hand surface locations. (b) Photo showing the gesture that elicited the wave patterns shown: tapping digit I to V on a flat surface consecutively. (c) Grabbing and lifting a handle. Wave propagation was reconstructed on a 3D hand model. (d) Time-averaged (\gls{rms} over a 250 ms window) vibration amplitude of tapping individual digits. (e) Five selected instants showing whole-hand tactile signal evolution during grabbing and lifting a handle.}
\label{fig:Reconstruct_Wave}
\end{figure*}

As explained in Section II, elasticity couples skin motion at different locations.  The spatial wavelengths of signals excited in the range of frequencies relevant to tactile perception are relatively long ($\lambda > 1$ cm), allowing a sparse Nyquist sampling approach.  Here, we demonstrate how to estimate the spatiotemporal skin motion across the hand from accelerometer measurements, by mapping them onto a standard 3D hand reconstruction.  Each accelerometer signal $\mathbf{a}_i(t)$ is associated with a  position $\pp_i$.  We associate these locations to the coordinates of the 3D hand model via anatomical registration of the anatomically-known sensor locations.  The hand model consists of a point cloud obtained from a high resolution 3D scanner (Model Eva, Artec3D). Using the model, we compute the geodesic distance $d(\pp_i,\pp)$ between sensor positions $\pp_i$ and arbitrary surface points $\pp$, using a shortest path algorithm.  We estimated the acceleration  $\mathbf{a}(\pp,t)$ at arbitrary hand surface points using a physiologically informed distance weighting (Fig.~11a)  \cite{manfredi2012effect, shao2016spatial}:
\begin{eqnarray}
\mathbf{a}(\pp,t) &=& \frac{\sum_{i=1}^{42}f(\phi(\pp,\pp_i))\mathbf{a}_i(t)}{\sum_{i=1}^{42}f(\phi(\pp,\pp_i))} \\
\phi(\pp,\pp_i) &=& \frac{17}{d(\pp,\pp_i)+\alpha}-C \label{eq:interpolation}\\
f(\phi) &=& \left\{ 
\begin{aligned}
\phi, \ \ \phi\geq 0\\
0, \ \ \phi< 0
\end{aligned}\right.
\end{eqnarray}
Here, $f(\phi)$ is half-wave rectification (replacing all negative values with zeros). This operation may be omitted if the signal phase is to be preserved. We evaluated these equations using values $\alpha$ = 25.5 mm and $C = 8.7\times10^{-2}$ that we selected based on previously published measurements \cite{manfredi2012effect}. $\phi(\pp,\pp_i)$ is used to compute the vibration amplitude at points $\pp$ based on measurement of sensor $\mathbf{a}_i(t)$ at $\pp_i$ accounting for damping with distance $d(\pp_i,\pp)$ (Fig.~\ref{fig:Reconstruct_Wave}(a)). 
While we could readily accommodate hand size, via a scale parameter $\gamma$, differences in hand shape and mechanics would require further steps. 

We applied this method in order to reconstruct spatiotemporal patterns of skin acceleration in the whole hand during manual interactions. Fig. 8  illustrates that vector acceleration signals $\mathbf{a}_i(t)$ recorded for all sensors and channels for one of these gestures, involving tapping the index finger, digit II.  Reconstructions of skin acceleration in the whole hand during manual interactions for an ensemble of different gestures are shown in Fig.~\ref{fig:Reconstruct_Wave}(d).  The figure depicts the peak instantaneous skin acceleration associated with each of five gestures that involved tapping each respective finger  (Fig.~\ref{fig:Reconstruct_Wave}(b)), revealing that these single-digit interactions elicited energy that readily propagated along the full extent of the respective digit, and into the rest of the hand. 

The wearable sensor captures tactile signals that are elicited in the whole hand during  object grasping and manipulation, as shown in Fig.~\ref{fig:Reconstruct_Wave}(c),(e). Here, grasping and lifting a handle excites skin vibration in all digits at different times, due to variations in contact timing. These vibrations readily propagate throughout the rest of the hand.

\section{Conclusion}\label{sec:conclusion} 
In this paper, inspired by human sensing, anatomy, and tissue biomechanics, we present a wearable tactile sensing array for large area remote vibration sensing in the whole hand.  The system comprises a 126-channel, wide bandwidth sensing array based on an ergonomically designed, flexible apparatus integrating 42 discrete three-axis accelerometers, and capable of capturing the propagation of vibrations, in the form of viscoelastic waves.  These waves propagate efficiently in hand tissues, distributing mechanical signatures of touch to widespread hand areas and sensory resources.  

The electronic design of our device is based on a custom FPGA based data acquisition system and sampling across 23 discrete I2C networks, ensures that the data across the array can be captured with high temporal fidelity and resolution, with an effective sampling rate of 1310 Hz. 
By carefully designing the physical configuration and ergonomics of this wearable sensor array we ensure that it can be used to accurately capture contact-elicited tactile signals during a variety of natural interactions, including touch contact, grasping, and manipulation of objects.  Experiments show that the individual sensors transduce skin acceleration with high temporal resolution.  We also showed that the device can accommodate a wide range of hand motions.  We also introduce physiologically informed methods for 3D reconstruction of the spatiotemporal propagation of tactile signals throughout the hand.   

Recent research has elucidates the remarkable abilities of the human hand to perform touch sensing at a distance, using remotely propagating viscoelastic waves that are excited via contact interactions with objects and surfaces.  The exquisite biological sensory apparatus of the human hand leverages these propagating tactile signals in order to enable a large variety of perceptual and manipulation tasks.  This research enables new methods for scientific inquiry, including studies of how tactile sensing in the whole hand relies on both mechanics and neural transduction to shaping tactile inputs that are the basis of touch perception and prehensile interaction.  It also provides a unique model for artificial touch sensing, in which contact interactions are encoded via a multitude of vibration sensors positioned remotely from the contact interface, and are synergetically combined through mechanical couping.  The instrument presented here also enables new methods for tactile sensing in wearable applications of interactive computing, and provides new quantitative methods for product design and usability that leverage these new tools for characterizing tactile signals that are felt during interaction with newly designed products or interfaces.  We anticipate that future devices inspired by the system presented here will facilitate applications of this approach in robotics, prosthetics, consumer electronics, wearable computing, augmented and virtual reality, health care, and many other applications.




\section*{Acknowledgment}
This work was supported by NSF awards No.~1446752, 1527709, 1751348, by a Google Faculty Research Award, and a Hellman Foundation Faculty Fellowship.

\ifCLASSOPTIONcaptionsoff
  \newpage
\fi



\bibliographystyle{IEEEtran}
\bibliography{IEEEabrv,shv2016wearable_bib}

\end{document}